\documentclass[10pt,twocolumn,letterpaper]{article}

\usepackage{cvpr}
\usepackage[accsupp]{axessibility}  
\definecolor{cvprblue}{rgb}{0.21,0.49,0.74}
\usepackage[pagebackref,breaklinks,colorlinks,allcolors=cvprblue]{hyperref}

\def\paperID{*****}
\def\confName{CVPR}
\def\confYear{2026}

\title{ViSAGE @ NTIRE 2026 Challenge on Video Saliency Prediction}
\author{Kun Wang$^1$ \quad
Yupeng Hu$^1$ \quad
Zhiran Li$^1$ \quad
Hao Liu$^1$ \quad
Qianlong Xiang$^{2,3,4}$ \quad
Liqiang Nie$^2$ \\
\\
$^1$Shandong University
$^2$Harbin Institute of Technology (Shenzhen) \\
$^3$City University of Hong Kong 
$^4$Shenzhen Loop Area Institute \\
{\tt\small \{khylon.kun.wang, zhiranli325, liuh90210, xiangqianlongcs, nieliqiang\}@gmail.com} \\
{\tt\small huyupeng@sdu.edu.cn}}
\begin{document}
\maketitle

\begin{abstract}
In this report, we present our champion solution for the NTIRE 2026 Challenge on Video Saliency Prediction held in conjunction with CVPR 2026. To exploit complementary inductive biases for video saliency, we propose \textbf{Vi}deo \textbf{S}aliency with \textbf{A}daptive \textbf{G}ated \textbf{E}xperts (ViSAGE), a multi-expert ensemble framework. Each specialized decoder performs adaptive gating and modulation to refine spatio-temporal features. The complementary predictions from different experts are then fused at inference. ViSAGE thereby aggregates diverse inductive biases to capture complex spatio-temporal saliency cues in videos. On the Private Test set, ViSAGE ranked first on two out of four evaluation metrics, and outperformed most competing solutions on the other two metrics, demonstrating its effectiveness and generalization ability. 
Our code has been released at \url{https://github.com/iLearn-Lab/CVPRW26-ViSAGE}.
\end{abstract}

\begin{figure*}[t]
    \centering
    \includegraphics[width=0.8\textwidth]{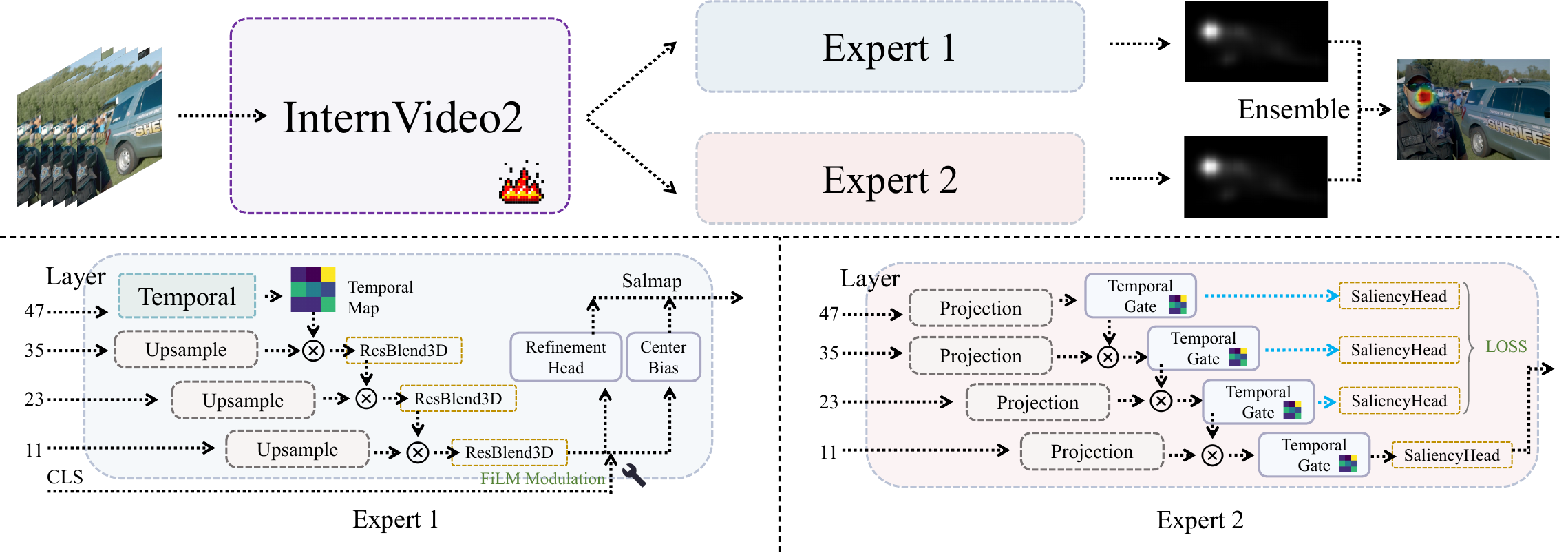}
    \vspace{-2mm}
    \caption{Overview of the proposed ViSAGE framework for video saliency prediction. Our method utilizes a shared InternVideo2 backbone to extract multi-scale spatio-temporal features, which are then processed by two complementary expert decoders: (1) a temporal-modulated decoder with spatial priors, and (2) a multi-scale decoder with deep auxiliary supervision. The outputs of both experts are fused at inference to generate the final saliency map, fully leveraging diverse inductive biases for robust and generalizable video saliency estimation.}
    \vspace{-4mm}
    \label{fig:methodology}
\end{figure*}

\section{Introduction}
Visual saliency prediction aims to estimate where humans are likely to attend in a visual scene, and has long served as a fundamental problem in computer vision, cognitive science, and multimedia analysis~\cite{itti1998model}. Compared with static image saliency, video saliency prediction is substantially more challenging~\cite{wang2018revisiting}, since human attention in videos is jointly influenced by spatial appearance, object semantics, motion patterns, temporal consistency, and scene dynamics~\cite{jiang2015salicon}. A reliable video saliency model therefore needs to capture not only fine-grained spatial structures, but also long-range temporal dependencies and cross-frame interactions. From a broader perspective, many studies on video understanding and moment-level reasoning share similar modeling demands~\cite{hu2021coarse,hu2023semantic,wang2024explicit,liu2025gaming,wang2026cross,zhang2024multi,hu2026visual}.

Recent advances in large-scale video representation learning have improved the modeling capacity for spatio-temporal understanding. Pretrained video backbones provide rich hierarchical features that benefit a wide range of downstream tasks~\cite{liu2022video,zhang2025spatial,internvideo2}. In parallel, lines of work on learning-based visual modeling, including temporal sequences and multimodal representation learning, continue to explore how dynamic visual signals are encoded~\cite{wang2021time,liu2025curmim,wang2025redundancy,zhang2023attribute}. Diffusion-based visual modeling has likewise drawn increasing attention, with studies on practical reuse and reliability~\cite{xiang2025dkdm,xiang2026tina,ho2022video}. Nevertheless, directly transferring such representations to video saliency prediction remains nontrivial. Saliency prediction differs from recognition-oriented tasks in that it requires dense, human-attention-aligned estimation, which is highly sensitive to multi-scale details, temporal transitions, and dataset-specific gaze biases~\cite{bylinskii2018different}. Existing methods often rely on a single decoding paradigm, such as top-down fusion, temporal attention, or multi-scale aggregation, which may capture only part of the complex factors that govern human visual attention. As a result, a single model often struggles to generalize across diverse video scenes with varying motion intensity, object layout, and saliency distribution~\cite{cheng2026towards}.

Representative video saliency models span diverse architectures, such as the temporally aggregating TASED-Net~\cite{min2019tased_net}, the audiovisual STAViS~\cite{tsiami2020stavis}, and the diffusion-based DiffSal~\cite{xiong2024diffsal}. A critical observation is that distinct architectural priors suit specific visual contexts. For instance, center-aware temporal reasoning effectively models stable gaze dynamics, whereas robust multi-scale fusion excels at resolving cluttered layouts, off-center targets, and multifocal salient regions. Motivated by these complementary strengths, our approach transcends the single-decoder paradigm by integrating specialized experts within a unified framework.

To address these challenges, we propose \textbf{Vi}deo \textbf{S}aliency with \textbf{A}daptive \textbf{G}ated \textbf{E}xperts (ViSAGE), a multi-expert ensemble framework designed to leverage complementary inductive biases for video saliency prediction. Instead of depending on a single decoder architecture, ViSAGE builds upon a shared InternVideo2 backbone~\cite{internvideo2} and introduces multiple specialized experts with distinct design principles. Concretely, one expert emphasizes temporally modulated decoding with explicit spatial priors, while another focuses on data-driven multi-scale fusion with deep auxiliary supervision. These experts produce complementary saliency estimates: the former is advantageous in scenes with stable temporal dynamics and strong center-oriented attention, whereas the latter is more flexible in handling complex layouts, off-center salient objects, and multi-region attention patterns. At inference, we average their saliency outputs for more robust modeling.

Our solution was developed for the NTIRE 2026 Challenge on Video Saliency Prediction~\cite{ntire26videosal}. The proposed framework ranked first on two out of four evaluation metrics on the Private Test set, and remained competitive on the remaining metrics, demonstrating effectiveness and generalization ability. Beyond challenge performance, ViSAGE also provides a simple and extensible recipe for combining multiple saliency experts on top of a pretrained video backbone.

In summary, our main contributions are threefold:
\begin{itemize}
    \item We propose ViSAGE, a multi-expert ensemble framework for video saliency prediction that exploits complementary inductive biases through specialized expert decoders.
    \item We design two structurally different saliency experts on top of a shared InternVideo2 backbone: one based on temporal modulation with spatial priors, and the other based on multi-scale fusion with deep auxiliary supervision.
    \item We demonstrate the effectiveness of the proposed framework on the NTIRE 2026 Video Saliency Prediction Challenge, where our method achieves leading results among submitted solutions.
\end{itemize}

\section{Methodology}
\label{sec:methodology}

We propose a multi-expert ensemble framework for video saliency prediction. As illustrated in Figure~\ref{fig:methodology}, our approach is built upon a shared InternVideo2 backbone~\cite{internvideo2} and incorporates two structurally complementary decoders, whose predictions are fused to generate the final saliency maps.

\subsection{Shared Backbone}
We adopt InternVideo2-Stage2\_6B as the spatio-temporal feature extractor. Multi-level representations are collected from four intermediate layers (layers 11, 23, 35, and 47), covering a spectrum from low-level spatial details to high-level semantic concepts. To adapt the pretrained backbone to the saliency prediction task while preserving its rich video representations, we fine-tune it using LoRA~\cite{lora}.

\subsection{Expert 1: Temporal-Modulated Decoder with Spatial Priors}
The first decoder derives an explicit temporal attention map from the deepest features (layer 47) and uses it to multiplicatively modulate shallower features in a top-down manner. At each stage, the modulated features are further refined using 3D residual blending blocks. In addition, Feature-wise Linear Modulation (FiLM)~\cite{film} is employed to inject global conditioning information. The final output is produced by a refinement head augmented with a learnable center-bias prior, which exploits the well-established tendency of human gaze to concentrate near the image center~\cite{tatler2007central}.

\subsection{Expert 2: Multi-Scale Decoder with Deep Auxiliary Supervision}
The second decoder first projects features from all four levels into a unified channel space, and then fuses them in a top-down manner through concatenation and 3D residual fusion blocks. A temporal gating mechanism is introduced at the deepest level to reweight features across time steps. Importantly, this decoder incorporates deep auxiliary supervision: intermediate saliency prediction heads are attached to multiple feature levels, providing dense supervisory signals that encourage each level to learn discriminative saliency-aware representations~\cite{xie2015holistically}.

\subsection{Complementarity and Ensemble}
The two experts provide complementary inductive biases. Expert 1 emphasizes multiplicative temporal modulation and explicit spatial priors, making it particularly effective in scenes with centrally located salient subjects and relatively stable temporal dynamics. In contrast, Expert 2 leverages concatenation-based multi-scale fusion and data-driven representation learning without relying on an explicit center prior, which improves adaptability to more complex scenarios involving off-center or multiple salient regions. The final saliency prediction is obtained by averaging the outputs of the two experts~\cite{ganaie2022ensemble}. Moreover, the proposed framework is naturally extensible: although our current implementation focuses on two experts, additional expert branches can be readily incorporated to accommodate other task-specific requirements in future extensions.

\subsection{Training Strategy}

We optimize each prediction head using a weighted combination of four standard saliency losses, including Kullback-Leibler (KL) divergence loss ($\mathcal{L}_{\mathrm{KL}}$), correlation coefficient (CC) loss ($\mathcal{L}_{\mathrm{CC}}$), similarity (SIM) loss ($\mathcal{L}_{\mathrm{SIM}}$), and binary cross-entropy (BCE) loss ($\mathcal{L}_{\mathrm{BCE}}$). Specifically, the KL divergence loss $\mathcal{L}_{\mathrm{KL}}$ measures the distribution discrepancy between the predicted saliency map and the ground-truth map after normalization. The CC loss $\mathcal{L}_{\mathrm{CC}}$ evaluates their linear correlation, and is implemented as one minus the correlation coefficient so that larger correlation leads to lower loss. The SIM loss $\mathcal{L}_{\mathrm{SIM}}$ measures the overlap between the normalized prediction and ground truth, and is defined as one minus their histogram intersection. The BCE loss $\mathcal{L}_{\mathrm{BCE}}$ is computed pixel-wise between the predicted saliency map and the target map. The overall training objective is defined as $10\mathcal{L}_{\mathrm{KL}} + 2\mathcal{L}_{\mathrm{CC}} + \mathcal{L}_{\mathrm{SIM}} + \mathcal{L}_{\mathrm{BCE}}$.

To elucidate the internal mechanisms of our specialized decoders, we briefly summarize their key components. We adopt a two stage projection with 3D convolutions to progressively compress backbone representations and align them into a unified channel space for top down fusion. To capture temporal dynamics, the Temporal Gate aggregates the deepest features through spatial global average pooling, followed by a one dimensional temporal convolution and a Sigmoid activation, producing a Temporal Map that is spatially broadcast to modulate shallower features multiplicatively. During fusion, adjacent multi scale features are concatenated and refined by ResBlend3D blocks, namely residual 3D convolutional modules with normalization and activation, to preserve fine grained spatio temporal details. In Expert 1, the FiLM conditioning signal is derived from the CLS token of the backbone’s final extraction layer and passed through two independent fully connected layers to predict affine modulation parameters. These layers are zero initialized to ensure an identity mapping at the beginning of training, thereby stabilizing early decoder learning.

\section{Experiments}

We report the performance of all methods on the NTIRE 2026 Video Saliency Prediction benchmark. The comparison is carried out according to four popular metrics for the saliency prediction task: CC, SIM, AUC Judd, and NSS. CC and SIM quantify the correlation and distributional similarity between predicted and ground-truth saliency maps; AUC Judd measures discrimination between fixated and non-fixated locations; NSS reports the average saliency at human fixation locations in units of the map's standard deviation~\cite{bylinskii2018different}. Higher values indicate better performance for all four metrics.

\subsection{Implementation Details}
We employ a two-stage training strategy. In the first stage, the backbone is frozen and only the decoder of each expert is trained, allowing the model to establish strong saliency decoding capability. In the second stage, LoRA modules are inserted into the backbone and jointly optimized with the corresponding decoder using a smaller learning rate. We optimize all trainable parameters with AdamW. Each expert is trained independently for 20 epochs under this two-stage protocol. We use a decoder learning rate of $1\times10^{-4}$ in Stage~1 and $1\times10^{-5}$ in Stage~2, while the LoRA parameters in Stage~2 are optimized with a learning rate of $1\times10^{-4}$. The batch sizes are 4 and 1 for Stage~1 and Stage~2, respectively. We retain the original 1080p challenge resolution for both landscape and portrait videos and uniformly sample 20 frames from each clip, where the clip length denotes the number of frames jointly processed by the model. Training is conducted on a single NVIDIA RTX PRO6000 GPU, and ensemble fusion is performed only during inference.

\begin{table}[t]
    \centering
    \caption{Cross-dataset evaluation on the public DIEM dataset. To evaluate broader generalization, our ViSAGE model is tested in a \textbf{zero-shot} setting (trained exclusively on the NTIRE split without any fine-tuning on DIEM).}
    \label{tab:generalization}
    \resizebox{\columnwidth}{!}{%
    \begin{tabular}{llcccc}
        \toprule
        Method & Venue & CC & SIM & AUC Judd & NSS \\
        \midrule
        ACLNet~\cite{wang2021revisiting_video_saliency} & TPAMI & 0.522 & 0.427 & 0.869 & 2.020 \\
        TASED-Net~\cite{min2019tased_net} & ICCV & 0.557 & 0.461 & 0.881 & 2.160 \\
        STAViS~\cite{tsiami2020stavis} & CVPR & 0.579 & 0.482 & 0.883 & 2.260 \\
        ViNet~\cite{jain2021vinet} & IROS & 0.632 & 0.498 & 0.899 & 2.530 \\
        TSFP-Net~\cite{chang2021temporal_spatial_feature_pyramid} & arXiv & 0.651 & 0.527 & 0.906 & 2.620 \\
        CASP-Net~\cite{xiong2023casp_net} & CVPR & 0.655 & 0.543 & 0.906 & 2.610 \\
        DiffSal~\cite{xiong2024diffsal} & CVPR & 0.660 & 0.543 & 0.907 & 2.650 \\
        \midrule
        \textbf{ViSAGE (Ours, Zero-Shot)} & \textbf{-} & \textbf{0.679} & \textbf{0.552} & \textbf{0.913} & \textbf{2.760} \\
        \bottomrule
    \end{tabular}%
    }
    \vspace{-2mm}
\end{table}

\subsection{Performance Comparison}

Table~\ref{tab:leaderboard} summarizes the private test scores officially reported by the challenge organizers~\cite{ntire26videosal}. Our method ranks first on CC and SIM; CVSP achieves the best AUC Judd, and ARK\_MMLAB attains the highest NSS. Overall, the top solutions are closely matched on localization-oriented metrics. The organizer-provided Baseline highlights the difficulty of the task and the large gap relative to learning-based methods.

To further assess the generalization capability of our approach and alleviate concerns about potential dataset-specific bias, we perform an additional evaluation on the widely-used DIEM dataset. Importantly, our final ViSAGE model is evaluated in a strict \textit{zero-shot} manner; that is, the model is trained only on the NTIRE split and no DIEM training data is used for fine-tuning.

As shown in Table~\ref{tab:generalization}, ViSAGE achieves robust performance. Under this zero-shot setting, our multi-expert ensemble attains state-of-the-art results on all four evaluation metrics, outperforming recent strong baselines such as CASP-Net and DiffSal. In particular, notable improvements are observed for NSS (2.760 vs. 2.650) and CC (0.679 vs. 0.660). These results suggest that our temporal-modulated and multi-scale experts learn complementary and generalizable representations of visual attention, enabling effective transfer to diverse real-world video scenarios.
\begin{table}[t]
    \centering
    \caption{Performance comparison on the NTIRE 2026 Video Saliency Prediction challenge using the official private test results. We report CC, SIM, AUC Judd, and NSS. Best results are in \textbf{bold}; second-best are \underline{underlined}.}
    \label{tab:leaderboard}
    \small
    \resizebox{0.8\columnwidth}{!}{%
    \setlength{\tabcolsep}{5pt}
    \begin{tabular}{lcccc}
        \toprule
        Team & CC & SIM & AUC Judd & NSS \\
        \midrule
        Baseline & 0.410 & 0.408 & 0.691 & 1.305 \\
        NTR & 0.693 & 0.579 & 0.845 & 2.989 \\
        SHU-MIPLab & 0.717 & 0.593 & 0.880 & 2.701 \\
        AAM & 0.748 & 0.579 & 0.815 & 3.100 \\
        Vertex & 0.796 & \underline{0.664} & 0.886 & 3.154 \\
        ARK\_MMLAB & 0.790 & 0.660 & 0.891 & \textbf{3.456} \\
        CVSP & \underline{0.827} & \underline{0.664} & \textbf{0.898} & \underline{3.416} \\
        \midrule
        \textbf{iLearn (Ours)} & \textbf{0.828} & \textbf{0.693} & \underline{0.892} & 3.323 \\
        \bottomrule
    \end{tabular}
    }
\end{table}

\subsection{Ablation Study}

\begin{table}[t]
    \centering
    \caption{Component ablation study evaluated on a held-out 10\% subset of the ECCV 2024 Video Saliency Prediction training data. We isolate the impact of core modules in each expert as well as the LoRA adaptation.}
    \label{tab:component_ablation}
    \small
    \setlength{\tabcolsep}{5pt}
    \scalebox{0.90}{%
    \begin{tabular}{lcccc}
        \toprule
        Model Variant & CC & SIM & AUC Judd & NSS \\
        \midrule
        \multicolumn{5}{l}{\textit{Expert 1 Variations}} \\
        Full Expert 1 & 0.794 & 0.634 & 0.916 & 3.225 \\
        \quad w/o center bias & 0.790 & 0.632 & 0.915 & 3.215 \\
        \quad w/o FiLM & 0.790 & 0.633 & 0.908 & 3.221 \\
        \quad w/o temporal modulation & 0.790 & 0.626 & 0.904 & 3.221 \\
        \quad w/o LoRA (Stage 1 only) & 0.783 & 0.624 & 0.914 & 3.180 \\
        \midrule
        \multicolumn{5}{l}{\textit{Expert 2 Variations}} \\
        Full Expert 2 & 0.794 & 0.636 & 0.916 & 3.220 \\
        \quad w/o temporal gate & 0.792 & 0.634 & 0.912 & 3.190 \\
        \quad w/o auxiliary heads & 0.790 & 0.635 & 0.910 & 3.220 \\
        \quad w/o LoRA (Stage 1 only) & 0.784 & 0.629 & 0.910 & 3.159 \\
        \bottomrule
    \end{tabular}
    }
\end{table}
To address the contribution of individual components, we conduct a granular ablation study on the held-out 10\% subset, as detailed in Table~\ref{tab:component_ablation}. For Expert 1, removing the temporal modulation leads to the most severe performance drop in SIM (from 0.634 to 0.626) and AUC Judd (from 0.916 to 0.904), highlighting the critical role of explicit temporal attention in capturing dynamic saliency shifts. The center bias and FiLM conditioning also provide consistent gains across all metrics. For Expert 2, the temporal gate is essential for fixating on dynamic salient regions, as its removal noticeably degrades the NSS score. The deep auxiliary heads contribute largely to spatial discriminability, improving AUC Judd from 0.910 to 0.916. Crucially, reverting both experts to Stage 1 (i.e., operating without LoRA fine-tuning on the backbone) results in the most significant overall degradation. This confirms that while the pretrained video backbone is powerful, adapting its representations via LoRA is indispensable for the dense, human-attention-aligned estimation required in this task.

To ensure objective evaluation, we reserve 10\% of the ECCV 2024 Video Saliency Prediction challenge training set~\cite{aim_challenge} as a held-out validation subset and report metrics under the same evaluation protocol. Our challenge submission fuses experts by equal averaging of their saliency outputs, as described in Section~\ref{sec:methodology}. Table~\ref{tab:ablation} additionally compares two-expert ensembles obtained by weighted fusion in logit space (followed by mapping back to saliency) versus weighted fusion directly on saliency maps, along with each expert. Both fusion strategies yield nearly identical overall accuracy: logit fusion provides a marginal SIM advantage ($+0.001$), whereas saliency fusion achieves the highest AUC Judd with the same CC score. We ultimately adopt logit-space fusion because it achieves the best SIM and slightly more consistent overall performance across metrics. Each single expert already delivers strong results: the fusion variants achieve higher CC, SIM, and NSS than either expert alone, while Expert~2 in isolation obtains a slightly higher AUC Judd, illustrating a mild trade-off across metrics rather than uniform gains. Overall, the two decoders maintain complementary behavior, and fusion leads to a more balanced multi-metric profile.

\begin{table}[t]
    \centering
    \caption{Ablation on ensemble design evaluated on a held-out 10\% subset of the ECCV 2024 Video Saliency Prediction training data. Best results are in \textbf{bold}; second-best are \underline{underlined}.}
    \label{tab:ablation}
    \footnotesize
    \resizebox{0.9\columnwidth}{!}{%
    \setlength{\tabcolsep}{3.5pt}
    \begin{tabular}{p{2.95cm}cccc}
        \toprule
        Variant & CC & SIM & AUC Judd & NSS \\
        \midrule
        Logit fusion (two experts) & \textbf{0.837} & \textbf{0.699} & 0.897 & \textbf{3.438} \\
        Saliency fusion (two experts) & \textbf{0.837} & \underline{0.698} & \underline{0.898} & \underline{3.420} \\
        Expert~2 only & 0.830 & 0.693 & \textbf{0.900} & 3.390 \\
        Expert~1 only & \underline{0.832} & 0.693 & 0.897 & 3.412 \\
        \bottomrule
    \end{tabular}
    }
\end{table}

\subsection{Case Analysis}

\begin{figure}[t]
    \centering
    \includegraphics[width=0.9\linewidth]{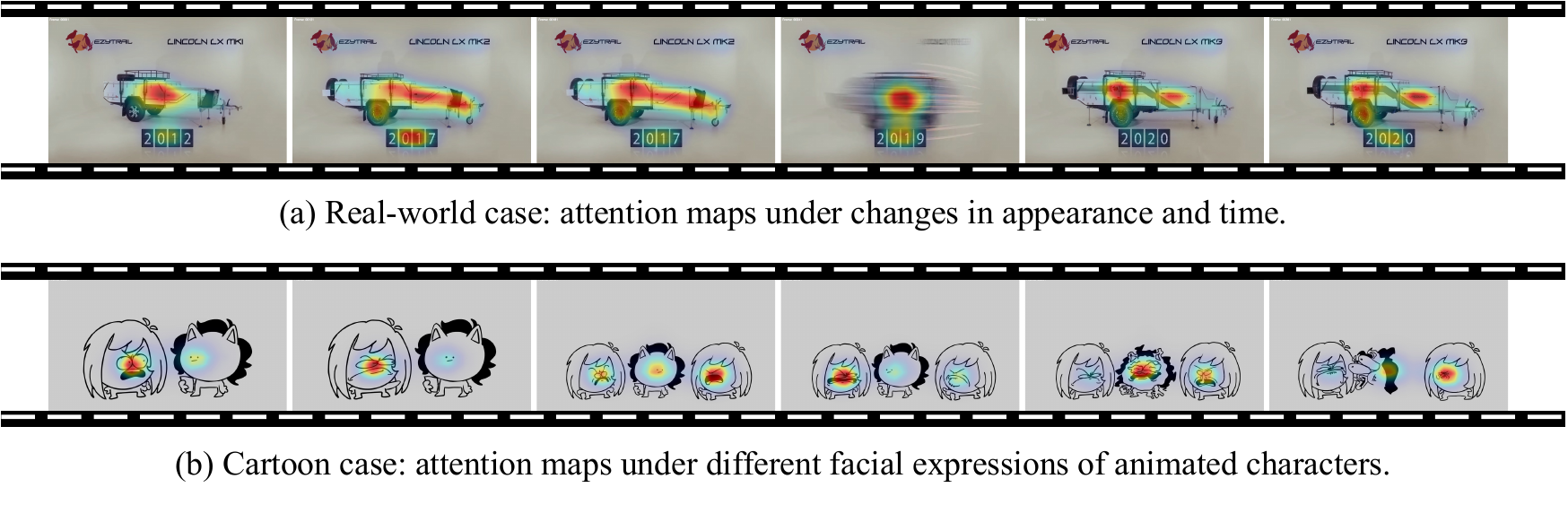}
    \vspace{-2mm}
    \caption{Visualization of predicted saliency under different types of variations.}
    \label{fig:attention}
    \vspace{-4mm}
\end{figure}

As shown in Figure~\ref{fig:attention}, we analyze two representative cases to better understand the model behavior under different visual variations. In the first case, which comes from real-world scenes, the target undergoes noticeable appearance and temporal changes, yet the model still concentrates on the semantically meaningful regions of the object, indicating a certain degree of robustness to real-world variation. In the second case, which involves animated characters, the visual content is dominated by changes in facial expressions. The predicted saliency maps show that the model is able to capture expression-related discriminative regions, suggesting that it can also adapt to more abstract and stylized visual patterns. These examples demonstrate that the model can respond to both realistic appearance/time variations and expression changes in cartoon scenarios, while also revealing how its focus shifts according to the underlying semantic cues of each case.

\section{Conclusion}

We present our champion solution to the NTIRE 2026 Video Saliency Prediction Challenge. ViSAGE is a multi-expert ensemble framework for video saliency prediction that leverages complementary inductive biases through specialized expert decoders. Specifically, we design two structurally distinct saliency experts on top of a shared InternVideo2 backbone: one based on temporal modulation with spatial priors, and another based on multi-scale fusion with deep auxiliary supervision. Our framework attains leading scores on two of the four official metrics while remaining competitive on the others, validating the effectiveness of fusing diverse expert decoders for robust and generalizable video saliency prediction.



{
    \small
    \bibliographystyle{ieeenat_fullname}
    \bibliography{main}
}
\end{document}